\documentclass[conference]{IEEEtran}
\IEEEoverridecommandlockouts
\RequirePackage{scrlfile} % Prevent svg (<2.0) package
\PreventPackageFromLoading{subfig}
\usepackage{cite}
\usepackage{amsmath,amssymb,amsfonts}
\usepackage{algorithmic}
\usepackage{graphicx}
\usepackage{textcomp}
\usepackage{xcolor}
\usepackage{subcaption}
\usepackage{verbatim}
\usepackage{tikz}
\usepackage{breqn}
\usepackage{booktabs}
\usepackage{multirow}
\usepackage{svg}
\usepackage{hyperref}
\usepackage[linesnumbered,ruled,vlined,algo2e]{algorithm2e}
\usepackage[inline]{enumitem}
\usepackage{balance}

\DeclareMathOperator*{\argmin}{arg\,min}

\def\BibTeX{{\rm B\kern-.05em{\sc i\kern-.025em b}\kern-.08em
    T\kern-.1667em\lower.7ex\hbox{E}\kern-.125emX}}
\begin{document}

\title{\LARGE \bf
PointNetKL: Deep Inference for GICP Covariance Estimation in Bathymetric SLAM}

% \author{Aleksandr Lyapunov \and Lev Pontryagin \and 
% \thanks{The authors are with the Division of Robotics, Perception and Learning at KTH Royal Institute of Technology, SE-100 44 Stockholm, Sweden
%         {\tt\small \{finn, jacke\}@kth.se}}%
% }

\author{
    \href{https://www.kth.se/profile/torroba}{Ignacio Torroba}\textsuperscript{$\star$} \\ 
    \and
    \href{https://www.kth.se/profile/sprague}{Christopher Iliffe Sprague}\textsuperscript{$\star$}\\
    \and
    \href{https://www.kth.se/profile/nbore}{Nils Bore} \\
    \and
    \href{https://www.kth.se/profile/johnf}{John Folkesson} \\
    \thanks{\textsuperscript{$\star$} Ignacio and Christopher contributed equally to this work.}
    \thanks{The authors are with the \href{https://smarc.se/}{Swedish Maritime Robotics Centre (SMaRC)} and the \href{https://www.kth.se/rpl/division-of-robotics-perception-and-learning}{Division of Robotics, Perception and Learning} at KTH Royal Institute of Technology, SE-100 44 Stockholm, Sweden.
    \{torroba, sprague, nbore, johnf\}@kth.se}
}

\maketitle

\begin{abstract}
Registration methods for point clouds have become a key component of many SLAM systems on autonomous vehicles.
However, an accurate estimate of the uncertainty of such registration is a key requirement to a consistent fusion of this kind of measurements in a SLAM filter. This estimate, which is normally given as a covariance in the transformation computed between point cloud reference frames, has been modelled following different approaches, among which the most accurate is considered to be the Monte Carlo method. However, a Monte Carlo approximation is cumbersome to use inside a time-critical application such as online SLAM. Efforts have been made to estimate this covariance via machine learning using carefully designed features to abstract the raw point clouds  \cite{landry2019cello}. However, the performance of this approach is sensitive to the features chosen. We argue that it is possible to learn the features along with the covariance by working with the raw data and thus we propose a new approach based on PointNet \cite{qi2017pointnet}. 
% PointNet is a recent neural net architecture designed to use arbitrarily sized and ordered point clouds as input.  
In this work, we train this network using the KL divergence between the learned uncertainty distribution and one computed by the Monte Carlo method as the loss.   
We test the performance of the general model presented applying it to our target use-case of SLAM with an autonomous underwater vehicle (AUV) restricted to the 2-dimensional registration of 3D bathymetric point clouds.
% demonstrate the how the model is able to learn the target distributions and show its usability within a SLAM framework, tested on two real missions with an autonomous underwater vehicle (AUV). 
% We evaluate this in terms of its generalizability using data from four different underwater locations ranging from relatively flat coastal areas to very rugged, deep Antarctic ocean areas. 
% To resolve these issues, we leverage recent advances in deep learning on point clouds to learn covariances directly from point clouds, circumventing the need to design features and saving on computational cost.
% We assess the performance of our method on real-world bathymetric data in both the regression and SLAM tasks.
\end{abstract}

\section{Introduction}
% It has been said that the best representation of sensor data for mapping is the data itself.  Even if this is not universally accepted, 
Over the last few years, sensors capable of providing dense representations of 3D environments as raw data, such as RGB-D cameras, LiDAR or multibeam sonar, have become popular in the SLAM community.  These sensors provide accurate models of the geometry of a scene in the form of sets of points, which allows for a dense representation of maps easy to visualise and render. 
The wide use of these sensors has given rise to the need for point cloud registration algorithms in the robotics and computer vision fields.
For instance, the core of well-established SLAM frameworks for ground and underwater robots, such as \cite{hess2016real} and \cite{teixeira2016underwater}, relies upon point cloud registration methods, such as the iterative closest point (ICP) \cite{besl1992method}, to provide measurement updates from dense 3D raw input.

While well rooted in indoor and outdoor robotics, SLAM has not yet gained widespread use for AUVs.  However, the need for SLAM is greater underwater, as navigation is extremely challenging.  Over long distances, sonar is the only viable sensor to correct dead-reckoning estimates.  Of the various types of sonar, multibeam echo sounders (MBES) provide the most suitable type of raw data for SLAM methods.  This data is essentially a point cloud sampled from the bathymetric surface, and applying registration methods to these measurements is a well-studied problem {\cite{torroba2018comparison}}.
% , one matches two patches of samples from seafloor to constrain the solution at all areas of overlap.  There are methods to find the best transformation between patches \cite{torroba2018comparison}.  
%The challenging part however is to
However, when fusing the output of the registration into the Bayesian estimate of the AUV state, the uncertainty of the transform must be modelled, since it represents the weight of the measurement.
Despite its importance in every SLAM domain, few works have addressed the problem of estimating this uncertainty accurately and efficiently, usually in the form of a covariance matrix. We argue that these two requirements are vital in our setting and need to be addressed simultaneously and onboard an AUV, with limited computational resources.
% Firstly, much of the uncertainty in the process is due to noise, low resolution in the sonar itself, and lack of features in the resulting bathymetric point cloud. Secondly, the solution must run onboard an AUV, with limited computational resources.
% Furthermore analytic solutions based on local properties of the geometry do not work as the registration is done over the whole submap and many local minima are typical.    

\begin{figure}[!t]
	\centering
	\includegraphics[width=0.45\textwidth]{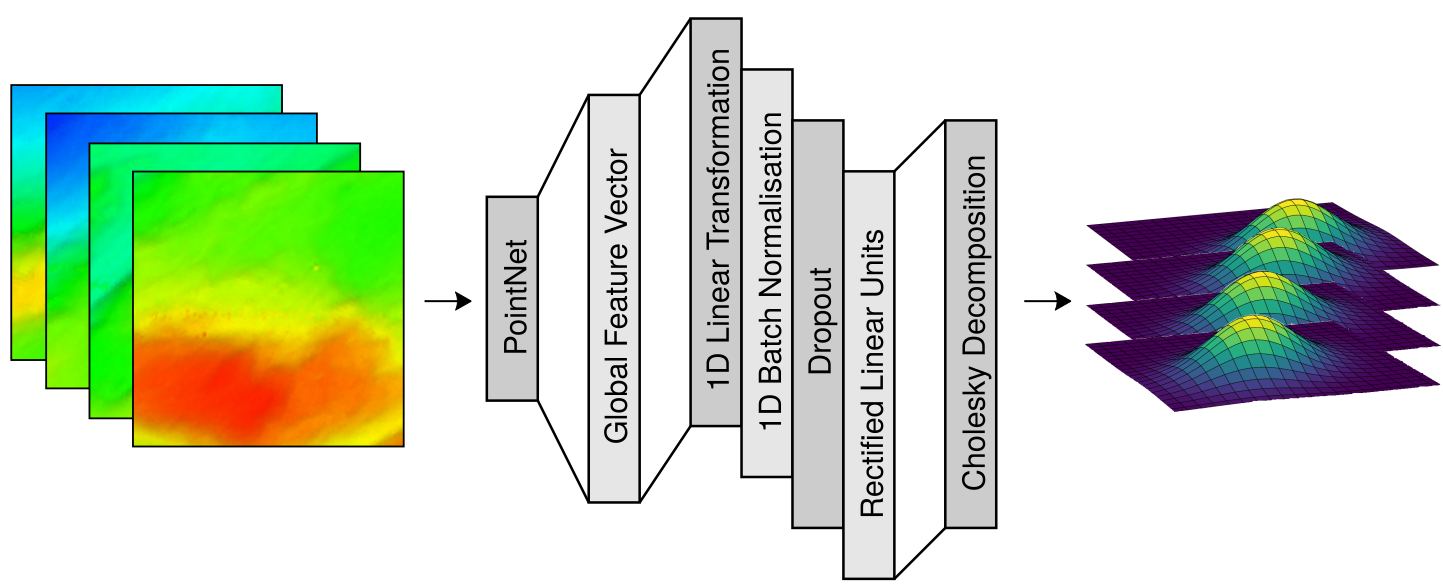}
	\caption{Depiction of the architecture of PointNetKL. From left to right:
	a submap is fed into PointNetKL to produce a global feature vector that is then fed into a multilayer perceptron to produce the parameters of a Cholesky decomposition and construct a positive-definite covariance matrix.
	}
	\label{fig:nn_architecture}
	\vspace{-0.3cm}
\end{figure}
% Accuracy in estimating the strength of the constraints from the measurements is crucial to robustly fusing all the information of the robot trajectory consistently in SLAM.  
% That is because the SLAM registration works by optimising  an objective function based on the trade-off between all information.  The fact that the patch to patch constraint is subjected to several sources of uncertainty which are hard to model and is by definition non-convex, complicates the computation of the uncertainty. 
% There exist methods to compute this probability distribution, ranging from assuming a constant covariance to Monte-Carlo estimation. 
% However, they are limited by their applicability, accuracy and/or computational cost.
% For this reason, we seek to develop a model of the underlying distribution of an ICP variant, the Generalized-ICP \cite{segal2009generalized}, that can provide a robust estimation in an online setting.
While there have been recent attempts to derive the covariance of the registration process analytically, these approaches were limited either by the reliability of the estimation or its complexity.
The latest and most successful techniques, however, have aimed at learning such a model.
\cite{landry2019cello} is an example of this approach applied to constraints created from point clouds registrations with ICP.
However, although successful, it is limited by the need to reduce the input point clouds to hand-crafted feature descriptors, whose design can be a non-trivial, task-dependent challenge.
Hence, we motivate our work with the goal to circumvent the need to design such descriptors, and to instead learn the features directly from the underlying raw data.
We accomplish this with the use of the relatively recent artificial neural network (ANN) architecture, PointNet \cite{qi2017pointnet}.
We combine PointNet and a parameterisation of a Cholesky decomposition of the covariance of the objective function into a single model, named PointNetKL, which is invariant to permutations of its input.
In this work, we use this architecture to estimate the Generalised-ICP (GICP) \cite{segal2009generalized} uncertainty distributions directly from raw data and test the learned model in a real underwater SLAM scenario.
Our contributions are listed as follows:

\begin{itemize}
    \item We present PointNetKL, a new learning architecture built upon PointNet, for learning multivariate probability distributions from unordered sets of $V$-dimensional points.
    \item We apply this general architecture to the restricted case of learning $2D$ covariances from the constrained GICP registration of real $3D$ bathymetric point clouds from several underwater environments.
    \item We assess the performance and generalisation of our architecture in both the regression and SLAM tasks.
\end{itemize}

\section{Formal Motivation}
% \subsection{Integration of GICP in SLAM}
\label{sec:motivation}
A point cloud registration algorithm can be defined as an optimisation problem aimed at minimising the distance between corresponding points from two partially overlapping sets of points, $S_{i}$ and $S_{j}$, with rigidly attached reference frames $T_{i}$ and $T_{j}$.
In the general case, a point cloud $P_i=\{S_i, T_i\}$ consists of an unordered set of $V$-dimensional points $S_{i}$ while a reference frame $T_{i} \in SE(3)$ represents a $6$-DOF pose.
According to this, an unbiased registration process can be modelled as a function $h$ of the form
\begin{equation}
\label{eq:gicp}
    T_{ij} = T^{-1}_{i} T_{j}= h(P_i,P_j),
\end{equation}
where  $h(P_i,P_j)$ is the true relative rigid transformation between the two frames.  This transformation is estimated to minimise the alignment error to $S_{j}$ when applied to $S_{i}$.
However, it is well known that due to the fact that point cloud registration locally optimises over a non-convex function, its solution is sensitive to convergence towards local minima and so its performance relies crucially on the initial relative transformation between the point clouds, computed as $\hat T^{-1}_{i} \hat T_{j} $.  Where the hat denotes the current best estimate of the pose. 

The uncertainty in the estimate of the true transformation can be represented within a SLAM framework as follows.
Given an autonomous mobile robot whose state over time is given by $x_{i} \in \mathbb{R}^{N}$ and with a dead reckoning (DR) system that follows a transition equation \ref{eq:slam_basic}, we can model a measurement update based on point cloud registration through Eq. \ref{eq:slam_2}, where $z_{ij} \in \mathbb{R}^{M}$ with $M \leq N$. 
\begin{align}
    x_{i} &= g(x_{i-1}, u_{i}) + \epsilon_{i} \label{eq:slam_basic} \\
    z_{ij} &= h(P_i, P_j) + \delta_{ij} \label{eq:slam_2}.
\end{align}

$\epsilon_{i}$ and $\delta_{ij}$ model the error in the DR and in the registration respectively. 
% Where $x_i$ are the parameterization of the $T_i$.  
Approximating the noise in the DR and measurement models as white Gaussian with covariances $R \in \mathbb{R}^{N \times N}$ and $Q \in \mathbb{R}^{M \times M}$, respectively, both $x_{i}$ and $z_{ij}$ will follow probability distributions given by 
\begin{align}
    x_{i} &\sim \mathcal{N}(g(x_{i-1}, u_{i}), R_{i}) \\
    z_{ij} &\sim \mathcal{N}(h(P_i, P_j), Q_{ij})
\end{align}
With $z_{ij}$ being measurements of the transform between point clouds from the registration algorithm and the true transform $h(P_i, P_j)$ being given by the $T_i$ and $T_j$. The Bayesian estimate that results from this model is generally analytically intractable due to the nonlinear terms, but iterative maximum likelihood estimates (MLE) are possible.
Such an iterative SLAM estimate requires a reliable approximation of the probability distribution of the registration error $\delta_{ij} \sim \mathcal{N}(0, Q_{ij})$.
This is because $Q_{ij}^{-1}$ represents the weight or "certainty" of the measurement $z_{ij}$ when being added to the state estimate, which is a critical step in any state-of-the-art SLAM solution. 
However, this covariance is not available and research has focused on deriving both analytical and data-driven methods to estimate it.
Over the next section, we revisit the most relevant of these methods and the previous work upon which our approach builds.

\section{Related work}
% existing registrations techniques - analytical 
As introduced above, the need of SLAM systems for a reliable approximation of the error distribution of point cloud registration processes has motivated a prolific body of work in this topic.
Among the numerous existing registration techniques, the ICP algorithm and its variants, such as GICP, are the most widely used in the SLAM community.
The methods to estimate the uncertainty of the solution of ICP-based techniques are traditionally divided into two categories: analytic and data-driven.
Analytical solutions based on the Hessian of the objective function, such as \cite{censi2007accurate}, have yielded successful results on the $2D$ case thanks to its capacity to model the sensor noise.
However, \cite{prakhya2015closed} shows that its extension to 3D contexts results in overly optimistic results, which do not reflect the original distribution. 
Another set of approaches, such as \cite{barczyk2017towards}, consists in developing estimation models for specific sensors. 
Although more accurate, this kind of method suffers in its inability to generalise to different sensors.

% estimation approaches that use hand-crafted features
There is a large body of work on non-parametric approaches to estimating probability distributions, e.g. \cite{buch2017prediction, tallavajhula2016nonparametric, landry2019cello}.
In \cite{buch2017prediction}, Iversen et al. apply a Monte-Carlo (MC) approach to estimate the value of the covariance of ICP on synthetic depth images. 
Although accurate, this approach cannot be applied online due to its computation time.
In \cite{tallavajhula2016nonparametric}, a general non-parametric noise model was proposed; however, its performance is adversely affected by scaling with higher-dimensional features.
Landry et al. introduce, in \cite{landry2019cello}, the use of the CELLO architecture for ICP covariance estimation.
This method proves to be a reliable estimation framework but it is limited by the need to create hand-crafted features from the raw 3D point clouds.
As argued in \cite{peretroukhin2015probe}, designing these features for a given task is still an open issue and so the success of CELLO is highly dependant on the chosen features.

% it's better to learn
Going beyond hand-crafted features, a number of works have addressed the problem of learning feature representations from unordered sets such as point clouds; however, these approaches do not capture spatial structure. As a result, there have been a number of point cloud representations developed through multi-view \cite{ savva2017large} and volumetric approaches \cite{qi2016volumetric}. Unfortunately, multi-view representations are not amenable to open-scene understanding, and volumetric representations are limited by data resolution and the computational cost of convolution.

The work in \cite{liu2018deep} presents an inference framework based on a deep ANN, DICE, that can be trained on raw images.
This work, to the best of our knowledge, represents the first instance of the use of a ANN to infer the uncertainty of a measurement model in a similar approach to ours.
However, their network is limited to camera input and they require ground truth measurements to construct the training set, a commodity often hard to afford.

% but we must learn in a good way
To counteract the need to preprocess the input point clouds as in \cite{landry2019cello} while being able to apply deep learning techniques as in \cite{liu2018deep}, we have turned to the seminal work PointNet \cite{qi2017pointnet} for our method. \textcolor{black}{When choosing between learning architectures, e.g. \cite{zhou2018voxelnet, wang2018dynamic}, we choose PointNet for its simplicity.  This being the first use of ANNs for bathymetric GICP covariance estimation, our results can serve as a first baseline.}

PointNet employs a relatively simple ANN architecture, that achieves striking performance in both classification and segmentation tasks upon raw point cloud data.
It relies on the principle of composing an input-invariant function through the composition of symmetric functions,
producing an input-invariant feature vector for a given point cloud.
% It abides to this principle by applying a shared multi-layer perceptron (MLP) and a set of transformations across all input points, then aggregating the outputs into a symmetric function, namely max-pooling.
% These operations produce a global feature vector for the whole point cloud and a feature vector for each point therein, both of which are invariant to permutations of the input point cloud.
% These learnt feature representations are then used further in the architecture for classification and segmentation.
% Recently, a number of further advancements to PointNet have been made \cite{qi2017pointnet++, shen2018mining, wang2018dynamic}.

While the PointNet architecture was originally intended for classification and segmentation, the internally generated feature vectors have been used for other purposes, such as point cloud registration \cite{aoki2019pointnetlk} and computation of point cloud saliency maps \cite{zheng2018learning}.
% In \cite{aoki2019pointnetlk} the architecture was used for point cloud registration.
% And in \cite{zheng2018learning} it was used to compute point cloud saliency maps, indicating which points have the most significant role in characterising an object's shape.
Similarly to these works, we seek to employ PointNet for a different purpose, namely for the estimation of multivariate probability distributions.

\section{Approach}
\label{seq:approach}
Our goal is to estimate, for each pair of overlapping bathymetric point clouds $P_{i}, P_{j}$, a covariance $Q_{ij}$ that is as close as possible to modelling the actual uncertainty of their GICP registration in underwater SLAM, given by $\delta_{ij} \sim \mathcal{N}(0, Q_{ij})$. 
Learning these covariances entails the use of datasets with large amounts of overlapping point clouds with accurate associated positions, such as the one used in \cite{landry2019cello}.
However, an equivalent dataset does not exist in the underwater robotics literature, due to the nature of bathymetric surveys with a MBES pointing downwards, where the consecutive MBES pings within swaths do not contain overlap.

In order to overcome this problem, we look into sources of uncertainty in the registration of bathymetric point clouds.
In general $Q_{ij}$ is a function of the amount of sensor noise, the statistics over the initial starting point for the iterative MLE, and the features in the overlapping sections of the two point clouds.
If we assume that $S_i$ has uniform terrain characteristics, i.e. there is not a small rocky corner on an otherwise flat point cloud, we can expect that $Q_i$ does not vary much with the specific region of overlap as long as that region is above a certain percentage of the whole point cloud.  
This allows us to attribute an intrinsic covariance to each point cloud independent of the actual overlap, i.e. $Q_{ij}$ becomes $Q_{i}$ (which is also approximately $Q_j$).   
The validity of this is based on the fact that it is possible to aggregate raw sonar data on fairly uniform point clouds with the submaps approach used in \cite{torroba2019towards}, for example.

\subsection{Learning architecture}
\label{subsec:cov_learning}
In this section we present a general approach to the problem of learning our target GICP covariance $Q_i$ from a set of points $S_{i}$.
Formally, we consider the problem of learning a function $\pi_\theta(S_i)$, parameterised by a set of learnable parameters $\theta$, that maps a point cloud of $U$ points $P_{i} = \{S_{i} \in \mathbb{R}^{U \times V}, T_{i} \in \mathbb{R}^{M}\}$ to a multivariate Gaussian probability distribution $\mathcal{N}\left(\mu_i, \Sigma_i \right) : \mu_i \in \mathbb{R}^M, \Sigma_i \in \mathbb{R}^{M\times M}$, where $\Sigma_i$ is strictly positive-definite.
We solve this problem by optimising $\pi_\theta$'s parameters $\theta$ to regress a dataset of the form 
\begin{equation}
\label{eq:dataset}
D=\left\{\left(S_{0}, \mu_{0}, \Sigma_{0}\right), \dots, \left(S_{K}, \mu_{K}, \Sigma_{K}\right)\right\},
\end{equation}
consisting of $K$  point clouds and their associated distributions.
% explain why we have zero-mean measurments?

%% write the rest of this when the sections are done...
% In sections \ref{subsec:cov_learning} to \ref{sec:klloss} we describe the design of the model used to learn $\pi_\theta$ and in section \ref{subsec:generation_training_data} we outline the generation of the training dataset in equation \ref{eq:dataset}.
% explain the architecture more
 
To tackle this problem we consider the PointNet architecture \cite{qi2017pointnet} in order to work directly with the raw point clouds.
We denote PointNet as function $\phi\left( S_i \right) : \mathbb{R}^{U \times V} \mapsto \mathbb{R}^{Z}$ that maps $S_i$ to a $Z$-dimensional vector-descriptor $\zeta_i \in \mathbb{R}^Z$, describing the features of $S_i$. 
% Internally, $\phi$ applies a series of MLPs and symmetric pooling functions to render $\zeta_i$ invariant to the order of $S_i$. 
% This is crucial as it is impractical to expect order standardisation when collecting raw sensor data.
Using $\zeta_i$ we seek to learn a further mapping to the distribution $\mathcal{N}\left(\mu_i, \Sigma_i \right)$. Given the invariance of $\zeta_i$, we postulate that the further abstraction to defining a probability distribution can be achieved by a simple MLP, denoted hereafter as $\psi(\zeta_i)$. 

In order to define our target distribution $\delta_{i}$, we task $\psi$ to output a covariance matrix $\Sigma_i = Q_{i}$ and set $\mu_i$ equal to the null vector.
% To define a multivariate probability distribution, we task $\psi$ to output a covariance matrix $\Sigma_i$, describing a zero-mean Gaussian distribution $\mathcal{N}\left(0, Q_i\right)$, corresponding to the form of the noise model $\delta_i$ introduced in Section \ref{sec:motivation}. 
We describe this model collectively as 
\begin{equation}
\pi_\theta(S_i) = \psi \left(\phi \left(S_i\right)\right) : \mathbb{R}^{U \times V} \mapsto \mathbb{R}^{M \times M},    
\end{equation}
where $\theta$ is the collective set of learnable parameters.
Hereafter, we denote $\pi_\theta$ as $\pi$ for brevity.

We leave the architecture of $\phi$ as originally described in \cite{qi2017pointnet}, removing the segmentation and classification modules. % and only outputting the global feature vector $\zeta_i$.
For the hidden model of $\psi$ we consider a fully connected feed-forward architecture of arbitrarily many layers and nodes per layer. 
For each layer we sequentially apply the following standard machine learning operations: 
linear transformation, 
$1D$ batch normalisation, 
dropout, 
and rectified linear units, 
as indicated in figure \ref{fig:nn_architecture}.
This characterisation of $\psi$ produces outputs in $\mathbb{R}_{\geq 0}$, which are transformed into desired ranges in the following section.

\subsection{Covariance matrix composition}

In order to map the outputs of $\psi$ to a valid estimation of $\Sigma_i$ we must enforce positive-definiteness. Following \cite{liu2018deep}, we task $\psi$ to produce the $(M^2 - M)/2 + M$ parameters of a Cholesky composition of the form
\begin{equation}
    \Sigma_i = L(l_i)D(d_i)L(l_i)^\intercal : l_i \in \mathbb{R}^{(M^2 - M)/2}, d_i \in \mathbb{R}_{>0}^{M},
\end{equation}
where $L(l_i)$ is a lower unitriangular matrix, $D(d_i)$ is a diagonal matrix with strictly positive values, and $[l_i,d_i]$ are the parameters to be produced by $\psi$. It is important to note that the strict positiveness of $D(d_i)$'s elements enforces the uniqueness of the decomposition and thus that of the probability distribution being estimated.
Using a linear transformation layer, we map the penultimate outputs of $\psi$ to $(M^2 - M)/2 + M$ values describing the elements of $l_i$ and $d_i$. To enforce the positivity of $d_i$, we simply apply the exponential function such that the decomposition becomes $\Sigma_i = L(l_i)D(\text{exp}(d_i))L(l_i)^\intercal$. We then use $\Sigma_i$ to fully characterise $\delta_i$.

% explain how to transform the outputs to \sigma after defining the loss function... because choosing LDL is good for computing the loss

\subsection{Loss function} \label{sec:klloss}
To train $\pi$, we must compare its predicted distributions with the true ones in order to compute its loss.
For this purpose we use the Kullback-Leibler (KL) divergence
\begin{multline}
    D_{KL}\left( \mathcal{N} || \mathcal{N}_\pi \right) = 
    \frac{1}{2} \biggl( 
        \text{tr}\left(
            \Sigma_\pi^{-1} \Sigma 
        \right)  \\ 
        + \left(\mu_\pi - \mu\right)^\intercal \Sigma_\pi^{-1} \left( \mu_\pi - \mu \right)
        - M 
        + \text{ln}\left( 
            \frac{\text{det}(\Sigma_\pi)}{\text{det}(\Sigma)}  
        \right) 
    \biggl),
    \label{Dkl}
\end{multline}
where $\text{tr}(\cdot)$ and $\text{det}(\cdot)$ are the trace and determinant matrix operations, respectively.
This gives us the distance between the distribution $\mathcal{N}(\mu_\pi, \Sigma_\pi)$ predicted by $\pi$ and the target distribution $\mathcal{N}(\mu, \Sigma)$.
We optimise $\pi$ to minimise $D_{KL}$, hence we coin its name, PointNetKL. As explained in \ref{subsec:cov_learning}, in our specific application $\mu=0$ and so the variable $\mu$ will be omitted for brevity from here on.

% \begin{figure*}[!t]
% 	\centering
% 	\includegraphics[width=\textwidth]{img/cannonical_sms_small.png}
% 	\caption{Canonical representation of the combined points distribution of the Baltic, Shetland and Antarctica datasets respectively.}
% 	\label{fig:canonical_datasets}
% \end{figure*}

% Note, these are .png with dpi=400 because the number of points is colossal

% \begin{figure}[!h]
% 	\centering
% 	\includegraphics[width=0.4\textwidth]{img/antarctica_cube.png}
% 	\caption{Before feeding the submaps $S_i$ to PointNetKL $\pi_\theta$, we sequentially zero-mean, normalise, and voxelise them with a size of $0.01$. Here we show a voxelised concatenation of a dataset.}
% 	\label{fig:canonical_datasets}
% % 	\vspace{-1cm}
% \end{figure}
 
\subsection{Generation of training data}
\label{subsec:generation_training_data}
% For training our network we would need to generate, for each pair of overlapping point clouds $P_{i}, P_{j}$, a $Q_{ij}$ that is as close as possible to modelling the actual pose uncertainty of their GICP registration in a SLAM task.  
% In general $Q_{ij}$ would be a function of the features in the overlapping sections of the two point clouds, the amount of sensor noise and the statistics over the initial starting point for the iterative MLE. As the true surface measured is usually unavailable, estimation of $Q$ must be based on the measured points $S_i$ and $S_j$. Furthermore, our method assumes that $Q$ does not vary much with the specific region of overlap as long as that region is above a certain percentage of the whole point cloud.  This allows us to attribute an intrinsic covariance to each point cloud independent of the actual overlap, i.e. $Q_{ij}$ becomes $Q_{i}$ (which is also approximately $Q_j$).   This assumption amounts to saying that $S$ have uniform terrain characteristics, i.e. there is not a small rocky corner on an otherwise flat point cloud.  The validity of this is based on the fact that it should be possible to aggregate raw sensor data on fairly uniform point clouds, although we did not do that as an explicit step with the data that we used for our evaluation.  These assumptions simplify the task and so are a good starting point.  

The datasets necessary to train, test and validate our ANN have been generated following an approach similar to \cite{buch2017prediction}. A Monte Carlo approximation has been computed for every point cloud $S_i$ as follows.
Given a 3D point cloud $P_i=\{S_i, T_i\}$ and a distribution $O \sim \mathcal{N}(0, \Sigma_{sample})$,  we generate a second point cloud $P_j$ by perturbing $P_i$ with a relative rigid transform $T_{j}$ drawn from $O$.
After the perturbation, Gaussian noise is applied to $S_j$ and the resulting point clouds are registered %according to equation \ref{}. %z_i equation.
using GICP.  Given the fact that $P_j$ is a perturbation of $P_i$, the error of the GICP registration can be computed as the distance between the obtained transformation $\hat T_j$ and the perturbation originally applied $T_j$, as in Eq. \ref{eq:error}, \cite{barfoot2014associating}.
This error is then used to calculate the covariance of the distribution $\delta_{i}$ for each point cloud following Eq. \ref{eq:cov}.
\begin{equation}
\label{eq:error}
    e_l = log(exp(\hat T_j)^{-1} T_j)
\end{equation}
\begin{equation}
\label{eq:cov}
    Q_i = \frac{1}{(L-1)} \sum_{l=1}^{L} e_l e^{T}_l
\end{equation}
Where $L$ is the total number of MC iterations per point cloud.

\section{Experiments}
In the remainder of this paper we apply the presented general approach for learning Gaussian distributions from unordered sets of points to the specific problem of bathymetric graph SLAM with GICP registration as introduced in \cite{torroba2019towards}. 
The reason to focus on GICP as opposed to ICP is that this method works better on the kind of bathymetric point clouds produced from  surveys of unstructured seabed, as discussed in \cite{torroba2018comparison}.

The nomenclature used this far can be instantiated to this specific context as follows.
A point cloud $P_i$ consists now of a bathymetric submap in the form of a point cloud $S_i \in R^{U \times 3}$ and the estimates of the AUV pose while collecting the submap are $T_i = exp(x_i) \in \mathbb{R}^{6}$.
In the case of the GICP registration, the vehicles used to collect the data provided a good direct measurement of the full orientation of the platform and the depth underwater. 
Due to this, the dimension of the measurement model $z_i$ in Eq. \ref{eq:slam} has been reduced to $m = 2$, since it is only the $x$ and $y$ coordinates that contain uncertainty.
Consequently, the GICP registration is constrained during the training data generation to the dimensions $x, y$ and therefore the covariances of $\delta_i$ and of the prior $O$ become $\mathbb{R}^{2}$ matrices.

\subsection{The training datasets} \label{subsec:trainingsets}
To the best of our knowledge, no dataset like the one in Eq. \ref{eq:dataset} with bathymetric submaps exists and therefore a new one has been created.
When designing such a dataset several criteria must be fulfilled for the training to be successful and having generalisation of the results in mind:
i) The success of GICP will be, to a certain degree, linked to the features in the submap being registered. 
In simple cases this can be easily interpreted by looking at the resulting covariance. 
Intuitively, on a perfectly flat submap, an elongated feature along the $y$ direction will ease the registration perpendicular to that axis, yielding low values of the covariance for the $x$ axis. Equivalently, the registration along $y$ will result in a bigger uncertainty since the submaps can slide along the feature. 
Thus, given that $\pi$ is learning a mapping from geometric features to covariance values, it is important that the dataset created contains enough variation in the bathymetry. 
ii) It is not possible to collect and train on enough data for $\pi$ to be able to generalise successfully to the whole sea floor worldwide. 
However, \cite{zhou2016terrain} proved that pieces of seabed can be successfully modelled through Gaussian processes.
Based on this it can be assumed that with a large and varied enough dataset, our model should be able to generalise to natural seabed environments never seen before.
This would help to circumvent the fact that this kind of data is very scarce and difficult to obtain in comparison to image or point cloud datasets.
iii) Different vehicles are used to collect the data as to ease generalisation.
iv) Ground truth (GT) is not available.

% Furthermore, different sensors and vehicle configurations during the survey
\begin{figure}[!h]
	\centering
	\includegraphics[width=0.45\textwidth]{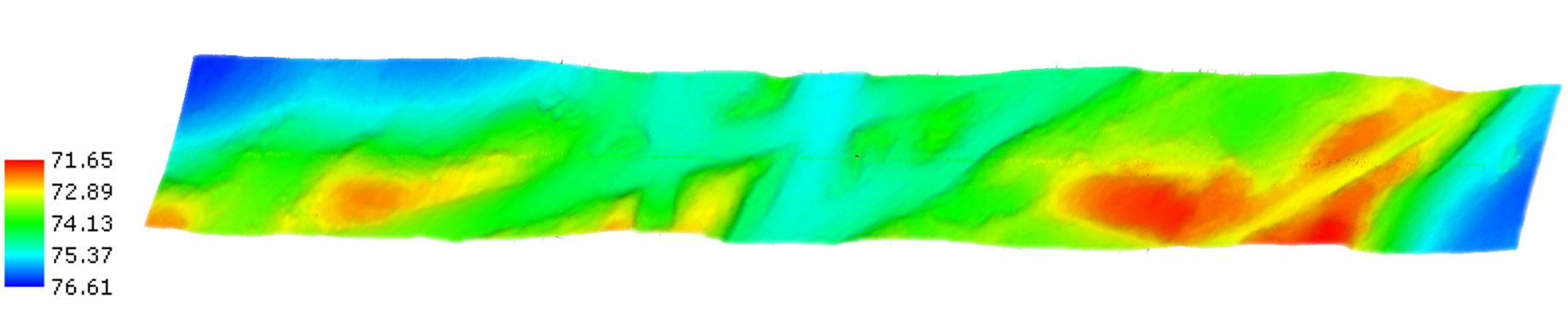}
	\caption{500 $\times$ 40 meters (approx.) sample of a swath of bathymetry from the Baltic dataset.}
	\label{fig:swath_bathy}
	\vspace{-0.4cm}
\end{figure}

\begin{figure}[ht]
    \centering
    \begin{subfigure}[b]{0.44\linewidth}
        \includegraphics[width=1\linewidth]{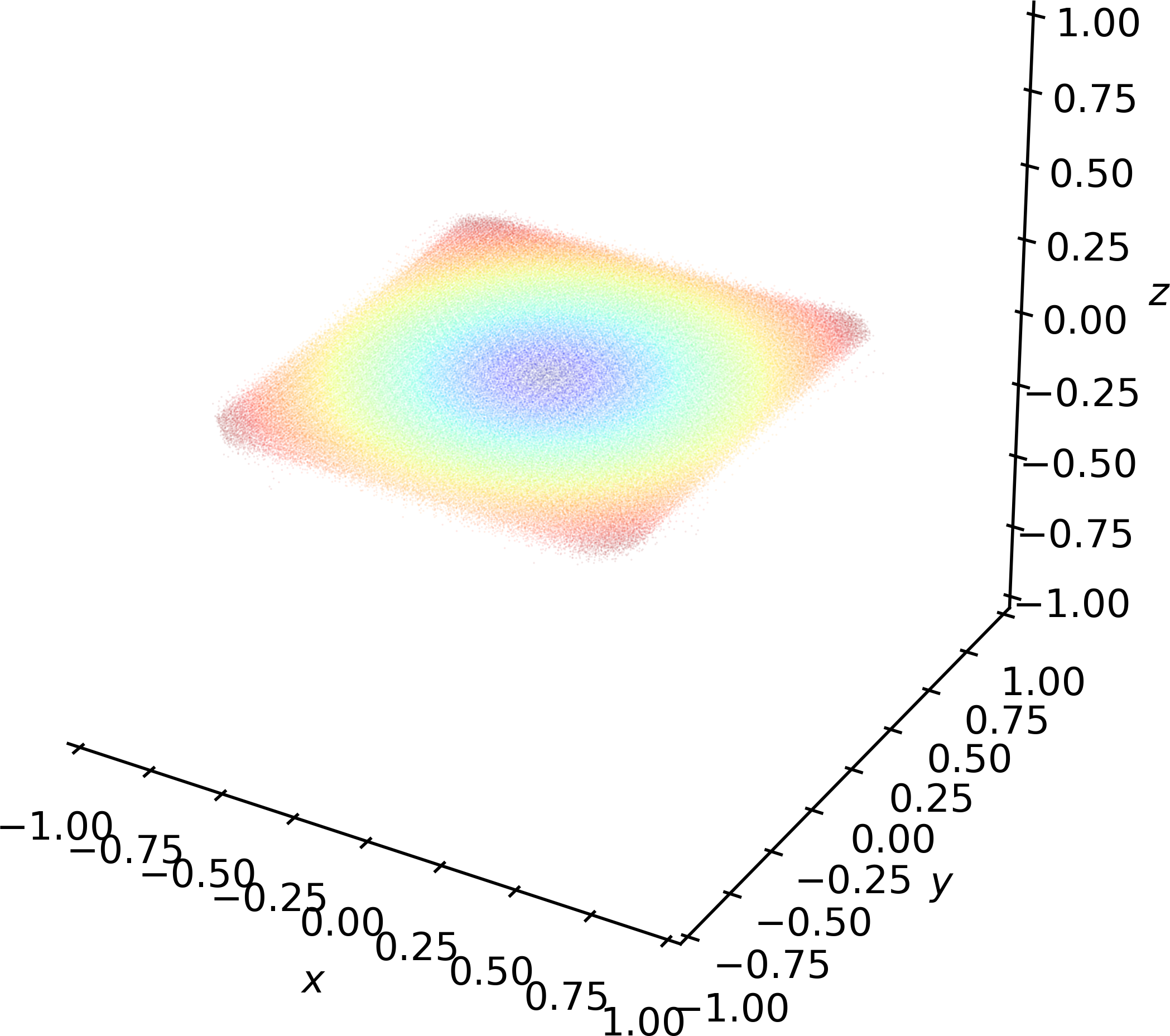}
        \caption{Baltic}
     \end{subfigure}
    \begin{subfigure}[b]{0.44\linewidth}
        \includegraphics[width=1\linewidth]{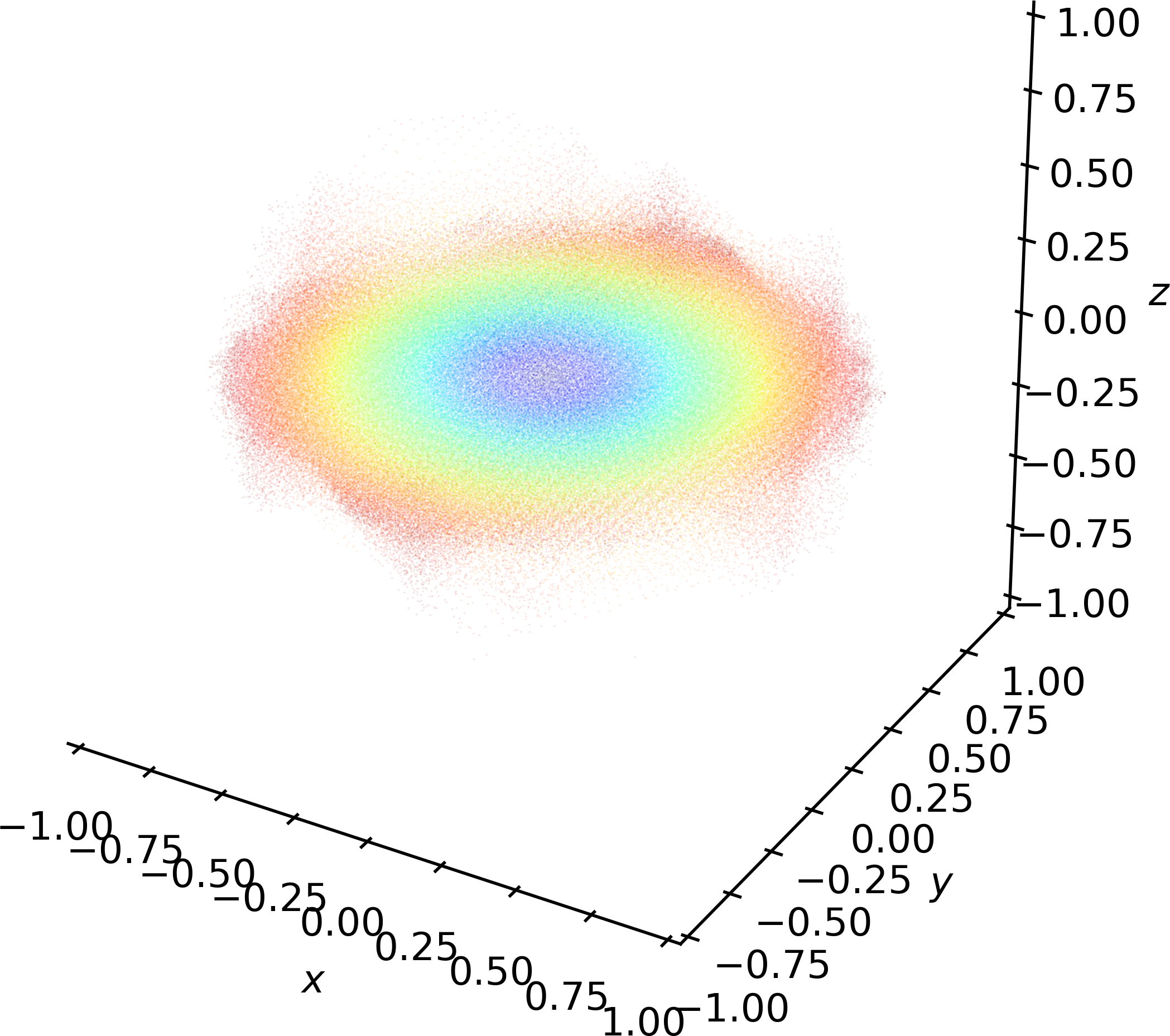}
        \caption{Born\"o 1}
     \end{subfigure} 
    % \begin{subfigure}[b]{0.24\linewidth}
    %     \includegraphics[width=1\linewidth]{img/borno_data.png}
    %     \caption{Born\"o}
    %  \end{subfigure} \hfill
    % \begin{subfigure}[b]{0.24\linewidth}
    %     \includegraphics[width=1\linewidth]{img/antarctica_data.png}
    %     \caption{Thwaites Glacier}
    %  \end{subfigure}

    \caption[]{
        % Before feeding the submaps $S_i$ to PointNetKL $\pi$, we sequentially , and voxelise them with a size of $0.01$. 
        Depiction of the zero-meaned, normalised, and voxelised submaps of two datasets.
        The dispersion of the points around the unit sphere indicate diversity of the bathymetry.
    }
    \label{fig:canonical_datasets}
\end{figure}

In order to ensure a complete distribution of geometric features within the dataset, we have analysed the spatial distribution of the points within the dataset used, with special attention to the dispersion in the $z$ axis, given by the standard deviation $\sigma_z$. 
Figure \ref{fig:canonical_datasets} shows a canonical representation of the zero-meaned, normalised, and voxelized submaps for two of the datasets used, whose characteristics are given in Table \ref{table:datasets}. 

% \begin{figure*}[ht]
%     \centering
%     \begin{subfigure}[b]{0.49\linewidth}
%         \includegraphics[width=1\linewidth]{img/borno_data.png}
%         \caption{Born\"o}
%      \end{subfigure} 
%     \begin{subfigure}[b]{0.49\linewidth}
%         \includegraphics[width=1\linewidth]{img/antarctica_data.png}
%         \caption{Thwaites Glacier}
%      \end{subfigure}

%     \caption[]{
%         % Before feeding the submaps $S_i$ to PointNetKL $\pi$, we sequentially , and voxelise them with a size of $0.01$. 
%         Depiction of the zero-meaned, normalised, and voxelised submaps of two of the datasets.
%         Note the dispersion of the points around the unit sphere, indicating the diversity of the bathymetry.
%     }
%     \label{fig:canonical_datasets}
% \end{figure*}

With the view on the criteria exposed above, four bathymetric surveys have been included in the final dataset.
They have been collected in four different environments, namely: the south-east Baltic sea, the Shetland Isles, Gullmarsfjorden near Born\"o, and underneath the Thwaites glacier in Antarctica.
A sample of bathymetry from the Baltic dataset can be seen in figure \ref{fig:swath_bathy}.
For the data collection, two state-of-the-art vehicles have been used. 
A remotely operated vehicle (ROV) Surveyor Interceptor and a Kongsberg Hugin AUV with an acoustic beacon Kongsberg cNode Maxiboth deployed in the survey area.
Both vehicles were equipped with a MBES EM2040.
With these four bathymetric surveys, the training datasets of the form given in Eq. \ref{eq:dataset} have been generated as explained in \ref{subsec:generation_training_data}, with $K= 9080$ submaps and $L=3000$ MC iterations per submap. 
Following the reasoning in \cite{landry2019cello}, we have set $\Sigma_{sample} = a I$ (where $I \in \mathbb{R}^2$) with $a = 9$ in order to model a reasonable underwater SLAM scenario.

\begin{table}[h]
\centering
\caption{Datasets characteristics}
\label{table:datasets}
\begin{center}
\begin{tabular}{|c|c|c|c|c|}
\hline
Area & Vehicle & Size ($km^{2}$) & Submaps & $\sigma_z$ $\in$ [0,1] \\
\hline
    Baltic   & Surveyor & 504.8 & 7543 &  0.067827 \\
    Shetland & Surveyor & 21.3 & 301 &  0.066160 \\
    Born\"o 1 & Hugin & 74.6 & 940   &  0.189505     \\
    Antarctica 7 & Hugin & 88.5 & 296 &  0.246802 \\
\hline
\end{tabular}
\end{center}
\vspace{-0.5cm}
\end{table}

% \begin{figure}[!h]
% 	\includegraphics[width=0.45\textwidth]{img/hugin_antarctica.jpeg}
% 	\caption{Kongsberg Hugin in the Thwaites glacier.}
% 	\label{fig:hugin}
% \end{figure}

% \begin{figure}[!h]
% 	\centering
% 	\includegraphics[width=0.45\textwidth]{img/canonical_pcls.png}
% 	\caption{Canonical representation of the points distribution of the submaps dataset.}
% 	\label{fig:pcls_canonical}
% \end{figure}

\subsection{Training implementation} \label{sec:training_details}
% Following the approach in Section \ref{subsec:generation_training_data}, a set of $K=1266$ submaps with their corresponding covariances has been created.
% Each submap has been registered $L = 3000$ times, with a perturbation distribution $O$ characterised by a $\Sigma_{sample} = a I$, where $I \in \mathbb{R}^2$ is the identity matrix.  
% Following the reasoning in \cite{landry2019cello}, we have set $a = 3$ meters in order to model a reasonable underwater SLAM scenario.

% architecture
In this work, we leave the internal architecture of PointNet $\phi$ as it is in \cite{qi2017pointnet}, outputting the vector-descriptor $\zeta \in \mathbb{R}^{1024}$.
For the MLP, mapping $\zeta$ to the estimation of GICP covariance $Q$, we consider an architecture of $4$ hidden layers, each having $1000$ nodes, using the sequential operations described at the end of Section \ref{subsec:cov_learning}.
A depiction of the final ANN can be seen in figure \ref{fig:nn_architecture}.

% training setup
To learn the mapping of submaps $S_i$ to covariances $Q_i$, we optimise the parameters of $\pi$ to regress the dataset $D$ under the cost function $D_{KL}$ given in Eq. (\ref{Dkl}). 
We optimise these parameters with stochastic gradient descent (SGD), using the AMSGrad variant of the adaptive optimiser, Adam \cite{kingma2014adam}.
In order to improve the generalisation of $\pi$ to different environments, we employ dropout \cite{srivastava2014dropout} and weight decay \cite{loshchilov2017fixing}.

For each training episode we sample a random subset (with replacement) of both the training and validation datasets, using the former for an optimisation iteration and the latter to evaluate the generalisation error in order to employ early-stopping \cite{yao2007early}.
We employ random subset sampling in conjunction with SGD in order to speed up training, as obtaining a gradient over the whole dataset described in Table \ref{table:datasets} is intractable.
% In this work we consider two separate instances of $\pi$, one trained and validated on the whole dataset (described in Section \ref{subsec:trainingsets}) and another excluding the data from Antartica. We refer to these two instance of PointNetKL as $\pi_\alpha$ and $\pi_\beta$, respectively.

The implemented training hyperparameters are:
\begin{enumerate*}[label=]
    \item learning rate $=1 \times 10^{-4}$,
    \item $L2$ weight decay penalty $=1 \times 10^{-4}$,
    \item dropout probability $=40\%$,
    \item batch size $= 500$,
    \item validation set proportion $=20\%$,
    \item early stopping patience\footnote{Number of iterations to wait after last validation loss improved.} $=20$.
\end{enumerate*}

Before the submaps are fed to network, they need to be pre-processed. Each set $S_i$ is translated to be zero-mean, then normalised to a sphere by the largest magnitude point therein, and finally voxelised to obtain a uniform density grid sampling.
This ensures that the raw density of each set $S_i$ does not affect the underlying relation that $\pi$ seeks to learn.
\textcolor{black}{Note, we use voxelisation here merely as a means to downsample the point clouds to lessen memory requirements. In principle, any downsampling method (not necessarily ordered) or none at all could be used.}

\subsection{Testing of the covariances in underwater SLAM}
\label{assess}
The validity of the covariances predicted by the network has been tested in the PoseSLAM framework in Eq. \ref{eq:slam}.
\begin{multline} 
\label{eq:slam} % this label doesn't work here
    \{x_i^*\} = \argmin_{\hat x} \sum_{i}^{N_{DR}} || g(\hat x_{i-1}, u_{i}) - \hat x_{i} ||^{2}_{R_{i}} \\
    + \sum_{\{i,j\}}^{N_{LC}} || \hat T^{-1}(\hat x_{i}) T(\hat x_{j}) - z_{ij} ||^{2}_{Q_{i} }
\end{multline}
Where $N_{DR}$ and $N_{LC}$ are the number of dead reckoning and loop closure (LC) constraints, respectively.
$Q_i$ models the weight of each LC edge added to the graph as a result of a successful GICP registration of overlapping submaps $P_i, P_j$ and as such it plays an important role in the optimisation.
% Refinements can be added if needed. 

For the tests, two underwater surveys outside the training dataset of the network have been used, named Born\"o $8$ and Thwaites $11$.
For the test on each scenario, a bathymetric pose graph is created and optimised similarly to \cite{torroba2019towards}.
The graph optimisations have been run with three different sets of covariances: those obtained with our method, the ones approximated with MC method and a constant covariance for each experiment.
The results of the optimisation processes have then been compared based on two different error metrics, $RMSE_{xyz}$, \cite{olson2009evaluating} and the map-to-map metric proposed in \cite{roman2006consistency}: 

\begin{itemize}
    \item $RMSE_{xyz}$: measures the error in reconstructing the AUV trajectory. 
    % However this alone is not enough as angular errors over long trajectories will yield a large ATE even when the trajectories are topologically similar but rotated.
    % ATE averages the RMS translation errors around the trajectories but proceeds this with a rigid transformation of on trajectory to minimise the leveraging effect of angular errors over long trajectories.  
    \item The map-to-map error: measures the geometric consistency of the final map on overlapping regions. 
    % Map-to-map error is a measure of the maximum minimum distance of points from one submap to points in another in overlapping regions.  Here the minimum is over points from the other submap. The maximum is over all the cells of a partition of the overlapping space, pairs of submaps and overlapping regions.  It measures the maximum 'thickness' of the combined point cloud, One could say the bathymetric resolution of the merged point cloud. 
\end{itemize}

The aim of these tests is to assess the influence of the GICP covariances on the quality of the PoseSLAM solution.
To this end, two modifications have been introduced in the construction of the pose graph with respect to \cite{torroba2019towards}: 
\begin{enumerate}
    \item The submaps created are all of roughly the same length.
    \item The initial map and vehicle trajectory are optimised in our graph SLAM framework using an estimate of the real $R$ from the vehicle and the MC estimated covariances for $Q$ in Eq. (\ref{eq:slam}) and used in lieu of actual ground truth, 'GT'.
\end{enumerate}

This second point can be further motivated by: i) the relatively high quality of the survey together with the absence of actual GT; ii) we will be disrupting the vehicle trajectory  by adding Gaussian noise much greater than the navigation errors.  Thus comparing the different optimisation outputs with respect to the undisrupted optimised estimate gives a valid comparison of methods as long as estimates are sufficiently further from the navigation than we believe the navigation is from the actual ground truth. More to the point we do not intend to prove that the MC method leads to a consistent estimate, for that we refer to the  \cite{buch2017prediction}.  Instead we wish to show that we can approximate well the solution that the MC method gives with our method. 

\begin{algorithm2e}[!h]
\small
    \caption{Corrupted pose graph construction}
    \label{algo:graph_construction}
    \SetKwFunction{FMain}{Create\_graph}
    \SetKwProg{Fn}{Function}{:}{}
    \Fn{\FMain{$\pmb{S_N}$}}{
        $\pmb{G} \leftarrow \{\emptyset\}$\\
        $\pmb{S_{graph}} \leftarrow \{\emptyset\}$\\
        \For{\texttt{$S_i$ in $\pmb{S_N}$}}{
            $\pmb{S_{LC}} \leftarrow \{\emptyset\}$\\
            $\pmb{G} \leftarrow \operatorname{AddDRedge}(S_i, S_{i-1})$ \\
            $\pmb{S_{LC}} \leftarrow \operatorname{DetectLC}(S_i, \pmb{S_{graph}}, coverage)$ \\
            \If{$\pmb{S_{LC}} \neq \{\emptyset\}$}{
                $\pmb{S_j} \leftarrow \operatorname{Perturb}(S_i, O)$ \\
                $\pmb{z_j} \leftarrow \operatorname{GICP}(\pmb{S_{LC}}, S_j)$ \\
                $\pmb{S'_i} \leftarrow \operatorname{Correct}(S_{j}, z_j)$ \\
                $\pmb{G} \leftarrow \operatorname{AddLCedge}(S'_i, \pmb{S_{LC}})$ \\
                $\pmb{S_{graph}} \leftarrow S'_{i}$\\
            }
            \Else{
                $\pmb{S_{graph}} \leftarrow S_{i}$\\
            }
        }
        $\pmb{G_{corrupted}} \leftarrow \operatorname{CorruptGraph}(\pmb{G}, R_c)$ \\
        \KwRet{$\pmb{G_{corrupted}}$}
    }
    % \label{alg:graph_construction}
\end{algorithm2e}
% \vspace{-0.4cm}

Algorithm \ref{algo:graph_construction} outlines the process followed to create the graphs for the optimisation tests.
As input it requires a 'GT' dataset, which we approximate optimising the navigation estimate from the vehicle DR as in \cite{torroba2019towards} using the MC covariances computed.
Given the estimated GT dataset divided into a set of $N$ submaps $\pmb{S_N}$, a graph $\pmb{G}$ is constructed in lines $2$ to $12$.
The initial bathymetry map from the undisrupted data can be seen in figures 5a and 5b.
In line $14$ the output graph is corrupted with additive Gaussian noise with covariance $R_c$.
An instance of the resulting bathymetry and graphs are shown in figures 5c and 5d respectively.
The arrows among consecutive submaps represent DR edges, while the non-consecutive ones depict the LC constraints.
The noise has been added to the graph once built instead of to the navigation data to ensure that the loop closure detections are preserved disregarding of the noise factor used.
However, an extra step must be taken to ensure a registration consistent with the assumptions on GICP initialisation as in section \ref{subsec:generation_training_data}.
In the case of a loop detection, the target submap $S_i$ is perturbed with a transformation drawn from $O$ before being registered against the fused submaps in $\pmb{S_{LC}}$.  That is we assume that there is, along with a good loop closure detection, an estimated starting point for GICP as given by the distribution $O$. 
The $coverage$ variable determines how much overlap must exist between two submaps for it to be considered a loop closure. It has been set to $60\%$ of $S_i$.
% Finally, if the registration is successful, LC edges are created between the registered $S_i'$ and the submaps in $\pmb{S_{LC}}$ and added to the graph.

% \begin{figure*}[!ht]
% \centering
% \begin{subfigure}[t]{.31\linewidth}
% 		\includegraphics[width=0.9\linewidth]{img/baseline_borno.pdf}
%     \caption{MC graph solution.}
% \label{subfig:baseline_result}
% \end{subfigure}
% \begin{subfigure}[t]{.31\linewidth}
% 		\includegraphics[width=0.9\linewidth]{img/nn_borno.pdf}
% 	\caption{NN graph solution.}
% \label{subfig:nn_result}
% \end{subfigure}
% \begin{subfigure}[t]{.31\linewidth}
% 		\includegraphics[width=0.9\linewidth]{img/fixedcov_borno.pdf} 
% 	\caption{Constant covariance graph solution.}
% \label{subfig:constant_result}
% \end{subfigure}
% \caption{Borno dataset during the covariance test through the PoseSLAM optimisation. The navigation estimate is in green, the corrupted path is red and the corrected in blue. }
% \label{fig:my_label}
% \end{figure*}

\begin{figure}[!h]
    \centering
    \includegraphics[width=0.9\linewidth]{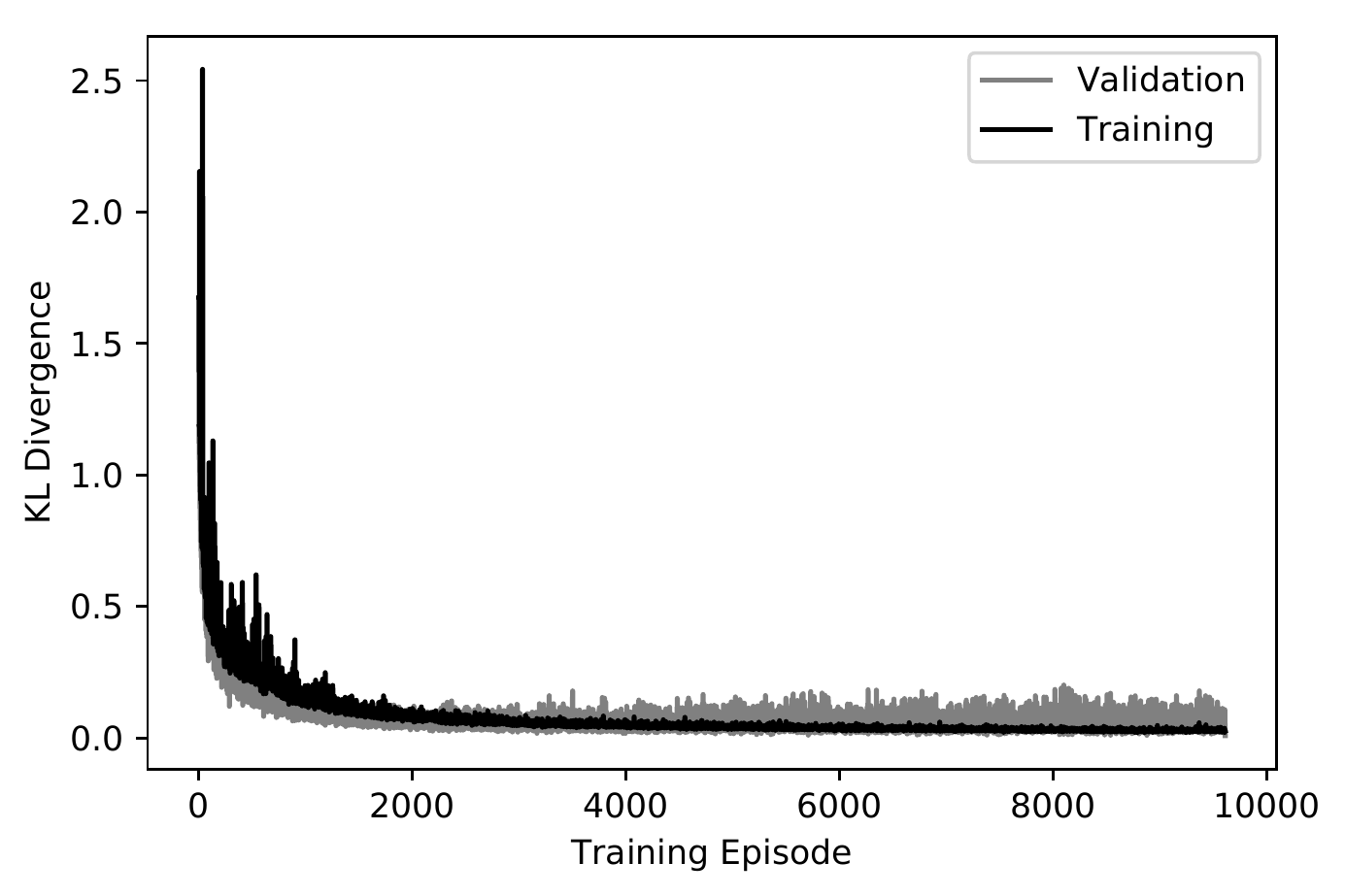}
    \caption{
    Evolution of $\pi$'s KL divergence loss on the training and validation sets.
    Note the unity of the training and validation curves over episodes, indicating generalisation.
    The jaggedness of the lines is a result of the stochasticity of the gradient descent, due to random subset sampling and dropout.
    }
    \label{fig:nn_training}
    \vspace{-0.4cm}
\end{figure}{}

\begin{figure*}[!ht]
\centering
\begin{subfigure}[t]{.24\linewidth}
\centering
	\includegraphics[width=0.9\linewidth]{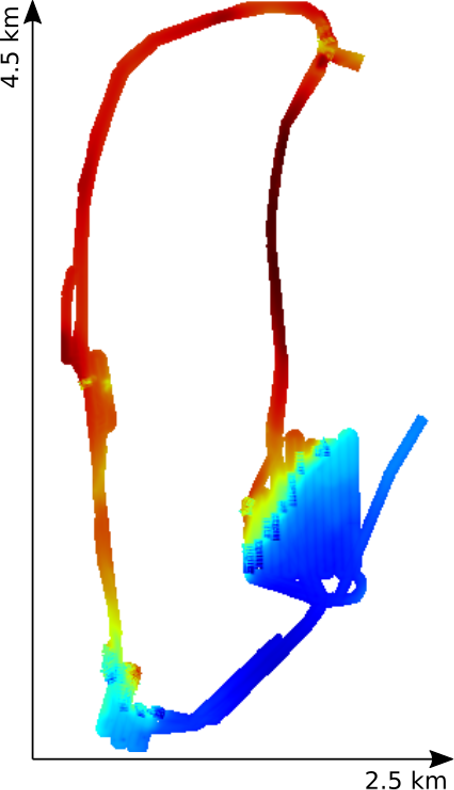}
%	\caption{Initial bathymetry map.}
\label{subfig:borno_bathy}
\end{subfigure}
\begin{subfigure}[t]{.24\linewidth}
\centering
        \includegraphics[width=0.81\linewidth]{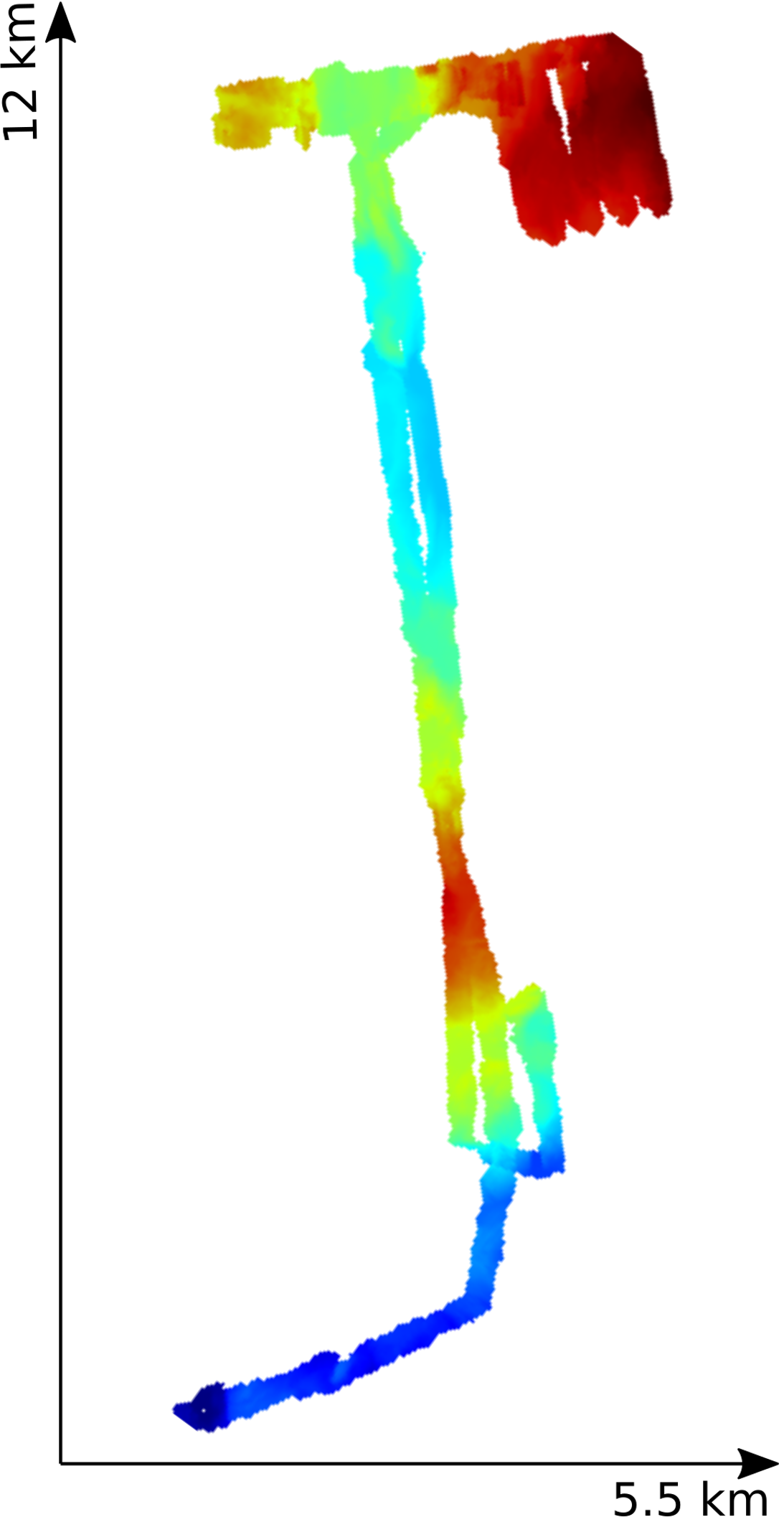}
 %   \caption{Corrupted dead reckoning.}
\label{subfig:antarctica_bathy}
\end{subfigure}
\begin{subfigure}[t]{.24\linewidth}
\centering
		\includegraphics[width=0.67\linewidth]{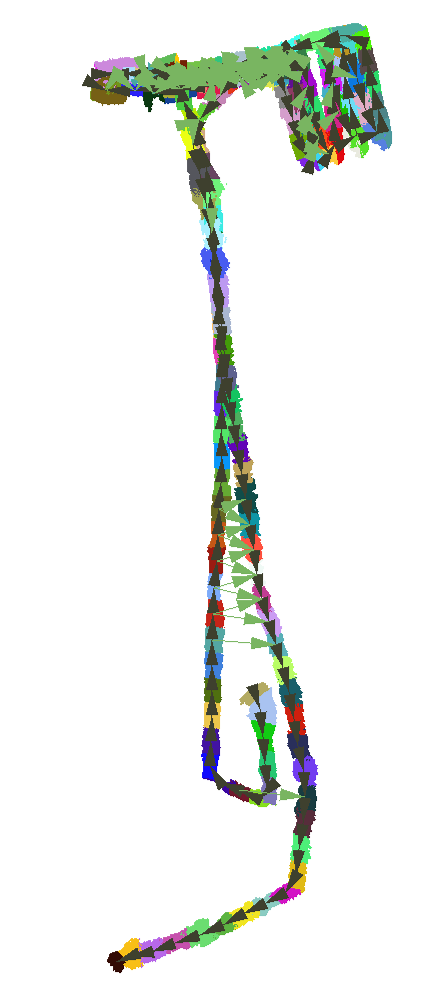}
%    \caption{Resulting graph}
\label{subfig:thwaites_graph_disp}
\end{subfigure}
\begin{subfigure}[t]{.24\linewidth}
\centering
		\includegraphics[width=0.67\linewidth]{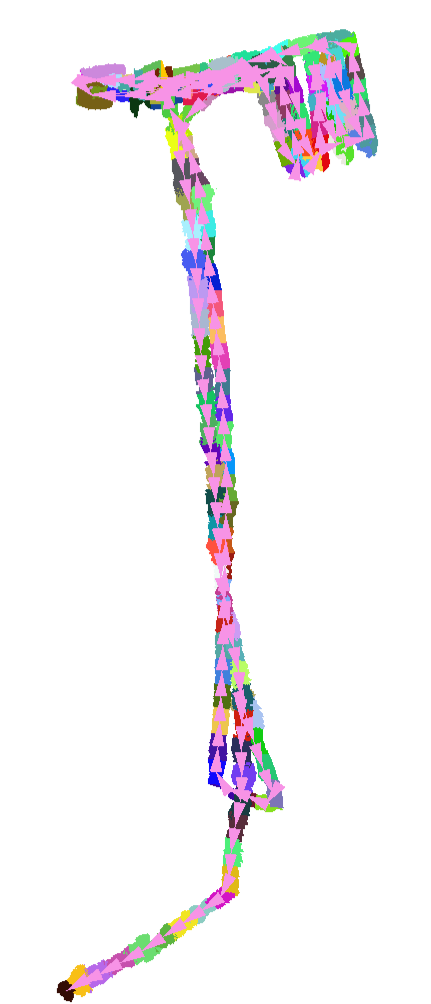}
  %  \caption{Resulting graph}
\label{subfig:thwaites_graph_fixed}
\end{subfigure}
\caption{Left to right: Born\"o $8$ and Thwaites $11$ surveys. Example of corrupted graph for the later ($RMSE_{xyz}$ 301.1 m) and its MC solution ($RMSE_{xyz}$ 222.18 m).}
\label{fig:slam_bathymetry}
\vspace{-0.4cm}
\end{figure*}

\section{Results}
\label{sec:results}
\subsection{Assessment of the NN predictions}
We assess the performance of $\pi$ on the training, validation, and testing sets with the KL divergence loss described in Section \ref{sec:klloss}.
In Figure \ref{fig:nn_training}, the training and validation loss maintain a strong coherence, indicating a good generalisation performance.
As indicated in table \ref{table:regression}, $\pi$ not only converges to low training and validation values, it also generalises well to the unseen testing sets.
Understandably, the network performs quite well in the Born\"o 8 validation set, in comparison to Thwaites 11, because the former is rather homogeneous in terms of feature types, whereas the latter includes many severe features on a larger scale. 
% Despite this, $\pi$ performs strongly in the SLAM task, as explained in the following section.

\begin{table}[h]
\centering
\caption{KL divergence loss of PointNetKL $\pi$ on the training and validation and testing sets.}
\label{table:regression}
\begin{center}
\begin{tabular}{|c|c|c|c|}
\hline
Training & Validation & Born\"o 8 & Thwaites 11 \\
\hline
    0.0184 & 0.0394 & 1.0578 & 1.9899 \\
    %$\pi_\beta$                & 0.0252 & 0.0412 & 0.3392 & 4.2333 \\
\hline
\end{tabular}
\end{center}
\vspace{-0.4cm}
\end{table}

\subsection{SLAM results}
The assessment of the covariances on an underwater SLAM context has been carried out on the two subsets from the Antarctica and Born\"o datasets in Figure \ref{fig:slam_bathymetry}. 
The results from the optimisation of the graphs generated from Algorithm \ref{algo:graph_construction} can be seen in Table \ref{tab:slam_results}. They have been averaged over $100$ repetitions where noise was added to the 'GT', which was then optimised. The $RMSE_{xyz}$ error under 'Navigation' indicates the correction from the DR estimate of the navigation after optimising it. \textcolor{black}{The resulting trajectory and bathymetry have been then corrupted with Gaussian noise parameterized by a covariance $R_c^{6x6}$ whose only non-null component is $R_c(5,5) = 0,01$, modelling the noise added to the yaw of the vehicle.}
The average errors from the corrupted graphs are given on the column "Corrupted".
The next four columns contain the final errors after the optimisations carried out with the covariances generated with the different methods tested.
\textcolor{black}{Our method has been compared against the covariances generated from the MC approximation in section \ref{subsec:generation_training_data} and against two baseline approaches.
A vanilla method consisting on approximating the information matrix of GICP, so called here "Na\"ive GICP", using the average of the information matrix in Eq. 2 in the original paper \cite{segal2009generalized}, projected onto the $xy$ plane for the current experiments.
This approximation doesn't require any extra computation and can be run during the mission, but is sensitive to the noise and sparsity of the point clouds.
For the second method, "Constant Q" as in \cite{liu2018deep}, all LC edges have the same $Q_i$, which in our case is computed a posteriori by averaging the MC solutions of all the submaps.}

When looking at the Thwaites $11$ results, considering the 'MC' method as the gold standard, it can be seen that it yields the smallest $RMSE_{xyz}$ error, as expected. Our method outputs a slightly worse $RMSE_{xyz}$ and similar Map-to-map error to MC and is significantly better than the two baseline approaches.
\textcolor{black}{However, in the Born\"o $8$ dataset our method's $RMSE_{xyz}$ error is better than the 'MC' output and very similar to the 'Na\"ive GICP'. This might be explained by the bathymetric data itself. The submaps from Born\"o contain far fewer features than those from Thwaites, and usually on the edges.  This makes the submap's terrain less homogeneous, violating our assumption on Section \ref{seq:approach}.  It is possible that the network had difficulty learning these non-homogeneous cases and tended to treat mostly flat submaps more like completely flat.  Then in the actual estimation, where the feature parts may not overlap at all, this resulted in a better output of the optimisation. 
The covariances from the 'Na\"ive GICP' are generally flat for large, noisy bathymetric point clouds.
This explains why they have performed so well in this dataset and so poorly in Thwaites 11, which contains more features.}

\textcolor{black}{\subsection{Runtime comparison}}
The execution times of 'MC' and 'Our' method have been compared on an Intel Core i7-7700HQ with 15,6 GiB RAM. 
The average covariance generation time over $107$ submaps of size $mean=6552.75, std\_dev=766.00$ can be seen in table \ref{table:timing}, supporting the claim that PointNetKL offers the best trade off between accuracy and processing time to run SLAM online on an AUV.

\begin{table}[!h]
\centering
\caption{Average runtime of the generation methods.}
\label{table:timing}
\begin{center}

\begin{tabular}{l|l|l|}
\cline{2-3}
                       & MC  & PointNetKL \\ \hline
\multicolumn{1}{|l|}{Runtime (s) $mean \pm std\_dev$} & 14.71 $\pm$ 5.08 & 0.13 $\pm$ 0.02 \\ \hline
\end{tabular}
\end{center}
\vspace{-0.4cm}
\end{table}

\begin{table*}[!t]
\captionsetup{size=footnotesize}
\caption{Graph-SLAM results for the four sets of covariances used in the optimisation of the two 'Corrupted' datasets. Observe how 'Our' method approximates well the 'MC' result for Thwaites 11 and even outperforms it for Born\"o 8.} \label{tab:slam_results}
\centering
\begin{tabular*}{0.99\textwidth}{@{}|lll|l|l|lllll|}
\hline
Dataset                      & Trajectory               & Graph constraints & Error (m)  & Navigation & Corrupted & Monte Carlo     & Constant Q & Na\"ive GICP & Ours   \\ \hline
\multirow{2}{*}{Thwaites 11} & \multirow{2}{*}{51.9 km} & $N_{DR}$ 222           & $RMSE_{xyz}$ \cite{olson2009evaluating}  & 15.3      & 323.4    & 248.1 & 287.1 &  302.8   & 259.5 \\ \cline{4-10} 
                             &                          & $N_{LC}$ 176           & Map-to-map \cite{roman2006consistency} & 32.43      & 34.17     & 33.75  & 34.04  &  33.75   & 33.65  \\ \hline
\multirow{2}{*}{Born\"o 8}     & \multirow{2}{*}{47.8 km} & $N_{DR}$ 395           & $RMSE_{xyz}$  \cite{olson2009evaluating} & 9.4       & 210.8    & 98.8  & 114.1    & 76.7 & 67.3  \\ \cline{4-10} 
                             &                          & $N_{LC}$ 301           & Map-to-map \cite{roman2006consistency} & 4.00       & 6.25      & 4.08   & 4.05  &  4.11  & 4.06   \\ \hline
\end{tabular*}
\end{table*}

\section{Conclusions}
We have presented PointNetKL, an ANN designed to learn the uncertainty distribution of the GICP registration process from unordered sets of multidimensional points.
In order to train it and test it, we have created a dataset consisting of bathymetric point clouds and their associated registration uncertainties out of underwater surveys, and we have demonstrated how the architecture presented is capable of learning the target distributions.
Furthermore, we have established the performance of our model within a SLAM framework in two large missions outside the training set.

The results presented indicate that PointNetKL is indeed able to learn the GICP covariances directly from raw point clouds and generalise to unknown environments. This motivates the possibility of using models trained in accessible environments, such as the Baltic, to enhance SLAM in unexplored environments, e.g. Antarctica. Furthermore, the data generation process introduced has proved to work well and alleviate the need for a dataset with sequences of overlapping point clouds with ground truth positioning associated, which are scarce in the underwater domain.
Further testing of the network on new graph-SLAM optimisations is required to fully characterise its performance and limitations, but the results presented support our thesis that PointNetKL can be successfully applied in online SLAM with AUVs. 

\textcolor{black}{The future extension of this work to the 3D domain will focus on estimating the heading of the AUV together with its $[x,y]$ state. 
Our intuition on this is that the same features in the submaps that anchor the registration in $[x,y]$ would lead the process if the GICP was unconstrained in the yaw as well.
% as far as the initial disruption fell within the convergence region.
This means that the features extracted by PointNetKL should, in general, perform well in 3D.}
% The exception to this would arise from cases equivalent to registering two perfectly spherical submaps.
% In this case, while $[x,y]$ would normally converge, the uncertainty on the relative orientation among submaps would be large.} 

% A tradeoff of the method is that if the assumption about the size of the odometry uncertainty made when creating the training datasets does not hold, the data generation and training steps need to be repeated. 

\section*{Acknowledgement}
The authors thank MMT for the data used in this work and the Knut and Alice Wallenberg foundation for funding MUST, Mobile Underwater System Tools, project that provided the Hugin AUV for these tests.
This work was supported by Stiftelsen fr Strategisk Forskning (SSF) through the Swedish Maritime Robotics Centre (SMaRC) (IRC15-0046).

\balance
\bibliography{main}
\bibliographystyle{unsrt}

\end{document}